\documentclass[journal]{IEEEtran}

\usepackage{lineno,hyperref}
\usepackage{multirow}
\modulolinenumbers[5]
\usepackage{amssymb} 
\usepackage{color}
\usepackage{amsmath,graphicx}
\usepackage{multirow}

\usepackage{graphicx,amsmath,bm}
\usepackage{xcolor}
\usepackage{subcaption}
\usepackage{algorithm}
\usepackage{algorithmic}
\usepackage{url}
\usepackage{multirow}

\def\words{{w}}

\def\ksentence{{k}}

\def\Cost{\cal{C}}
\def\C{{\Cost}}
\def\Cnorm{\C^{norm}}
\def\sub-words{{tokens}}
\def\sub-word{{token}}
\def\SP{{SentencePiece}}
\def\unigram{Unigram}

\def\ttwo{0.101}

\def\tone{-716.87}
\def\tonep{716.87}
\def\tzero{2.47* 10^{6}}
\def\dtwo{6.89 * 10^{-5}}

\def\done{0.24}
\def\dzero{21.23}
\def\tnorm{t^{norm}}
\def\dcoef#1{d_{#1}}
\def\tcoef#1{f_{#1}}

\def\ddcoef#1{d_{#1}}
\def\ttcoef#1{f_{#1}}

\def\dccoef#1{g_{#1}}
\def\tccoef#1{h_{#1}}

\def\ddtwo{2.48 * 10^{-8}}
\def\ddone{-1.76*10^{-4}}
\def\ddzero{3.06*10^{-3}}

\def\tttwo{2.37 * 10^{-8}}
\def\ttone{8.37*10^{-5}}
\def\ttzero{-3.40*10^{-3}}

\def\Deltaexp{\Delta_{exp}}
\def\Thetaexp{\Theta_{exp}}

\def\exp{e}

\begin{document}
\title{A Calculus-Based Framework for Determining Vocabulary Size in End-to-End ASR}

\author{Sunil Kumar Kopparapu\\ TCS Research - Mumbai.}

\maketitle

\begin{abstract}
In hybrid automatic speech recognition (ASR) systems, the vocabulary size is unambiguous, typically determined by the number of phones, bi-phones, or tri-phones present in the language. In contrast, end-to-end ASR systems derive their vocabulary, often referred to as \sub-words from the text corpus used for training. The choice and, more importantly, the size of this vocabulary is a critical hyper-parameter in training end-to-end ASR systems.
Tokenization algorithms such as Byte Pair Encoding (BPE), WordPiece, and Unigram Language Model (ULM) use the vocabulary size as an input hyper-parameter to generate the sub-words employed during ASR training. Popular toolkits like ESPNet provide a fixed vocabulary size in their training recipes, but there is little documentation or discussion in the literature regarding how these values are determined. Recent work \cite{kopparapu2024cost} has formalized an approach to identify the vocabulary size best suited for end-to-end ASR, introducing a cost function framework that treats the tokenization process as a black box.
In this paper, we build upon that foundation by curve fitting the training data and using the  principle of first and second derivative
tests in calculus to formally estimate the vocabulary size hyper-parameter. We demonstrate the utility and usefulness of our approach by applying it on a standard Librispeech corpus and show that the optimal choice of vocabulary size hyper-parameter improves the performance of the ASR.
The main contribution of this paper in formalizing an approach to identify the vocabulary size best suited for training an end-to-end ASR system.

\end{abstract}


\section{Introduction}
\label{sec:introduction}

End-to-end automatic speech recognition (ASR) systems have become the dominant paradigm for speech-to-text modeling by directly mapping acoustic feature sequences to discrete sub-word or character sequences. While this approach eliminates explicit phonetic and lexicon modeling, it introduces several representation-level design choices that significantly affect model performance. One such choice is the vocabulary size used in sub-word tokenization, which determines the resolution of the discrete output space employed during training and decoding.

In contrast to hybrid ASR systems, where the vocabulary is implicitly fixed by the phonetic inventory of the language, end-to-end ASR systems rely on data-driven tokenization algorithms such as Byte Pair Encoding (BPE), WordPiece, and the Unigram Language Model. These tokenization methods require the vocabulary size to be specified \emph{a priori}. Popular ASR tool-kits, including ESPNet, typically adopt fixed heuristic values for this parameter; however, the rationale behind these choices is often undocumented, and the relationship between corpus statistics and the optimal vocabulary size remains poorly understood.

Recent work~\cite{kopparapu2024cost} addressed this issue by introducing a cost-minimization framework that models the effect of vocabulary size through corpus-derived statistics while treating the tokenizer as a black box. Although effective, the proposed approach relies on empirical grid search over candidate vocabulary sizes and does not explicitly characterize the analytical structure of the resulting cost function.

In this paper, we extend this line of work by developing a calculus-based framework for estimating the vocabulary size in end-to-end ASR systems. We view vocabulary size as a \emph{representation resolution parameter}, analogous to dictionary size in vector quantization or model order in statistical signal processing. By explicitly modeling corpus-dependent cost components as smooth, differentiable functions of the vocabulary size, we derive first- and second-order optimality conditions that enable principled estimation of the optimal vocabulary size without exhaustive experimentation.

Specifically, we incorporate normalization of cost components based on corpus statistics to ensure numerical stability and interpretability across datasets. We further employ improved curve-fitting techniques using second-order polynomials and polynomial--exponential models to capture the observed behavior of token imbalance and sequence length growth as functions of vocabulary size. The resulting formulation allows the optimal vocabulary size to be obtained by solving a constrained stationary-point problem, providing analytical insight into how corpus characteristics govern tokenizer design.

Experimental validation on the LibriSpeech-100 corpus using a state-of-the-art conformer-based ASR model demonstrates that vocabulary sizes estimated by the proposed framework achieve competitive or improved word error rates compared to commonly used heuristic choices. These results highlight the utility of analytically guided representation design in end-to-end ASR systems. The main contributions of this paper are summarized as follows:

\begin{itemize}
    \item We formulate vocabulary size selection in end-to-end ASR as a continuous optimization problem by modeling corpus-derived cost components as differentiable functions of vocabulary size.
    
    \item We derive first- and second-order optimality conditions that characterize the existence of an optimal vocabulary size, enabling analytical or numerical estimation without exhaustive grid search.
    
    \item We introduce normalization of cost components based on corpus statistics, allowing stable optimization and meaningful interpretation of cost weights across datasets.
    
    \item We empirically validate the proposed framework on the LibriSpeech-100 corpus using a conformer-based ASR system and demonstrate competitive or improved word error rates relative to widely adopted heuristic vocabulary sizes.
    
\end{itemize}

The remainder of this paper is organized as follows. Section~\ref{sec:problem} introduces the extended problem formulation and analytical framework. Section~\ref{sec:experiments} presents the experimental setup and evaluates the proposed approach on a standard ASR benchmark. Section~\ref{sec:conclude} concludes the paper and discusses limitations and future directions.

\section{Problem Setup}
\label{sec:problem}

We adopt the problem formulation and notation from \cite{kopparapu2024cost}, where the vocabulary size is determined by minimizing a cost function that jointly accounts for (i) the number of tokens, (ii) class imbalance, and (iii) computational cost.
Specifically, the cost function comprises three terms: the vocabulary size $n$, the ratio of frequent to infrequent token occurrences $\Delta(n)$, and the total number of tokens required to cover the corpus $\Theta(n)$.
For completeness, we restate the cost function $\C(n)$ from \cite{kopparapu2024cost} as
\begin{equation}
     \C(n) =  \left \{ \alpha_1 \overbrace{n}^{t_1} + 
     \alpha_2 \overbrace{\Delta(n) }^{t_2} + 
     \alpha_3 \overbrace{\Theta(n)}^{t_3} \right \}
     \label{eq:mimimize}
\end{equation}
where $\alpha_{1,2,3}$ denote the weights associated with the respective components $t_{1,2,3}$. 
The optimal vocabulary size corresponds to
\begin{equation}
   n^* = \arg\min_{n} \left \{ \C(n) \right \} 
   \label{eq:nstar}
\end{equation}

In this work, we extend the formulation by normalizing the cost components to ensure stability and comparability across corpora. The cost terms are modeled using second-order polynomials and polynomial–exponential forms to accurately capture their empirical behavior. The optimal vocabulary size is obtained analytically by solving the first- and second-order optimality conditions. This provides a principled mechanism for estimating the vocabulary-size hyper-parameter in end-to-end ASR systems, which constitutes the main contribution of this work.

\subsection{Finding optimal number of \sub-words\ (\texorpdfstring{$n$}{n})}

We first compute the first derivative of $\C(n)$ with respect to $n$:
\begin{equation}
    \frac{d\C(n)}{dn} =
    \alpha_1 +
    \alpha_2 \Delta'(n) +
    \alpha_3 \Theta'(n),
    \label{eq:derivative}
\end{equation}
and set it to zero, yielding
\begin{equation}
    \alpha_1 +
    \alpha_2 \Delta'(n) +
    \alpha_3 \Theta'(n) = 0.
    \label{eq:derivative_min}
\end{equation}
We assume that the first derivatives $\Delta'(n)$ and $\Theta'(n)$ exist. Solving~\eqref{eq:derivative_min} yields a candidate solution for $n$, whose value depends on the weighting coefficients $\alpha_{1,2,3}$.

To verify that this solution corresponds to a minimum, we compute the second derivative
\begin{equation}
    \frac{d^2\C(n)}{dn^2} =
    \alpha_2 \Delta''(n) +
    \alpha_3 \Theta''(n),
    \label{eq:doublederivative}
\end{equation}
assuming the existence of $\Delta''(n)$ and $\Theta''(n)$. The solution $n$ obtained from~\eqref{eq:derivative_min} minimizes $\C(n)$ if~\eqref{eq:doublederivative} is strictly positive.

\subsection{Polynomial-Based Functional Modeling}
    The functions $\Delta(n)$ and $\Theta(n)$ are modeled using (i) a second-order polynomial and 
(ii) a second-order polynomial augmented with an exponential term, respectively. 
This choice ensures the existence of well-defined first- and second-order derivatives while providing sufficient flexibility to capture empirical trends.

\subsubsection{Second-Order Polynomial Fit}
\label{sec:second_order_fit}

We first model $\Delta(n)$ and $\Theta(n)$ using second-order polynomials:
\begin{equation}
\Delta(n) \triangleq \dcoef{2}n^2 + \dcoef{1}n + \dcoef{0},
\end{equation}
\begin{equation}
\Theta(n) \triangleq \tcoef{2}n^2 + \tcoef{1}n + \tcoef{0}.
\end{equation}
The corresponding first- and second-order derivatives follow directly:
\begin{equation}
\Delta'(n) = 2\dcoef{2}n + \dcoef{1}, \quad
\Theta'(n) = 2\tcoef{2}n + \tcoef{1},
\end{equation}
\begin{equation}
\Delta''(n) = 2\dcoef{2}, \quad
\Theta''(n) = 2\tcoef{2}.
\end{equation}
Substituting these into~\eqref{eq:derivative_min} yields
\begin{equation}
n =
\frac{-(\alpha_1 + \alpha_2\dcoef{1} + \alpha_3\tcoef{1})}
{2(\alpha_2\dcoef{2} + \alpha_3\tcoef{2})}.
\label{eq:found_generic_n}
\end{equation}
The solution corresponds to a minimum if  $\frac{d^2\C}{dn^2}>0$, namely
\begin{equation}
2(\alpha_2\dcoef{2} + \alpha_3\tcoef{2}) > 0,
\label{eq:generic_n_min_condition}
\end{equation}
and yields a positive $n$ provided (numerator in (\ref{eq:found_generic_n}) is negative)
\begin{equation}
\alpha_1 + \alpha_2\dcoef{1} + \alpha_3\tcoef{1} < 0.
\label{eq:generic_n_num_condition}
\end{equation}
Note that 
$n$ 
is dependent on the values of $\alpha_{1,2,3}$ which need to be chosen heuristically.
\subsubsection{Polynomial--Exponential Model}
\label{sec:second_order_plus_fit}

Empirical observations (see Section~\ref{sec:sopf}) indicate that a pure second-order polynomial does not adequately model the behavior of $\Delta(n)$ and $\Theta(n)$, both of which exhibit exponential-like trends. To address this, we augment the polynomial model with an exponential term:
\begin{equation}
\Delta_{\mathrm{exp}}(n) \triangleq
\dccoef{3}n^2 + \dccoef{2}n +
\dccoef{1} e^{1/n} + \dccoef{0},
\label{eq:deltac}
\end{equation}
\begin{equation}
\Theta_{\mathrm{exp}}(n) \triangleq
\tccoef{3}n^2 + \tccoef{2}n +
\tccoef{1} e^{1/n} + \tccoef{0}.
\label{eq:thetac}
\end{equation}
Solving the resulting nonlinear equation numerically under the constraint
\begin{align}
    \frac{d^2\C(n)}{dn^2} &=
    \alpha_2 \!\left( 2\dccoef{3} +
    3 \dccoef{1}\frac{e^{1/n}}{n^4} \right) \nonumber\\
    &+
    \alpha_3 \!\left( 2\tccoef{3} +
    3 \tccoef{1}\frac{e^{1/n}}{n^4} \right) > 0
    \label{eq:generic_cn_min_condition}
\end{align}
yields the optimal vocabulary size (see Appendix \ref{sec:second_order_plus_fit_a}).

\section{Experimental Analysis}
\label{sec:experiments}

To estimate the optimal vocabulary size $n$ following the methodology described in the previous section, we first model $\Delta(n)$ and $\Theta(n)$ and \textit{select} the weighting coefficients $\alpha_{1,2,3}$ to construct the cost function in~\eqref{eq:mimimize}. 
All experiments are conducted on the LibriSpeech-100 corpus~\cite{train-clean-100}. 
LibriSpeech-100 consists of 100 hours of read English speech along with corresponding text transcriptions. 
The training set contains $\ksentence=28{,}538$ sentences comprising $\words=990{,}093$ word tokens, of which $\words_u=33{,}798$ are unique. 
The corpus contains a total of $c=5{,}298{,}301$ characters, with $c_u=28$ unique characters.

The LibriSpeech-100 dataset is chosen for two reasons: (i) it is one of the most widely used benchmarks for training low-resource end-to-end ASR systems, and (ii) it is supported by a well-established training recipe in the ESPNet toolkit, which we employ in our experiments.

We compute $\Delta(n)$ and $\Theta(n)$ for values of $n$ in the range $c_u \leq n \leq 5000$ using the LibriSpeech-100 corpus. 
Figure~\ref{fig:delta} illustrates the behavior of $\Delta(n)$, while Figure~\ref{fig:theta} shows the corresponding $\Theta(n)$ curve, both derived directly from the corpus statistics.

\subsection{Second Order Polynomial Fit}
\label{sec:sopf}

We now fit a second-order polynomial (as described in Section~\ref{sec:second_order_fit}) using the {\tt curve\_fit} function from the {\tt scipy.optimize} module in the {\tt SciPy} library.
The estimated coefficients for $\Delta(n)$ are
$\dcoef{2}= \dtwo$,
$\dcoef{1}= \done$,
and $\dcoef{0}= \dzero$, with an $R^2$ value of $1.00$.
Similarly, for $\Theta(n)$, we obtain
$\tcoef{2}= \ttwo$,
$\tcoef{1}= \tone$,
and $\tcoef{0}= \tzero$, with an $R^2$ value of $0.73$.
The fitted models are given by
\begin{eqnarray}
    \Delta(n) & = & \dtwo n^2 + \done n + \dzero \nonumber \\
    \Theta(n) & = & \ttwo n^2 + \tone n + \tzero
    \label{eq:polyfit}
\end{eqnarray}

\begin{figure}[!htb]
\centering
\begin{subfigure}{0.495\textwidth}
    \includegraphics[width=.8\columnwidth]{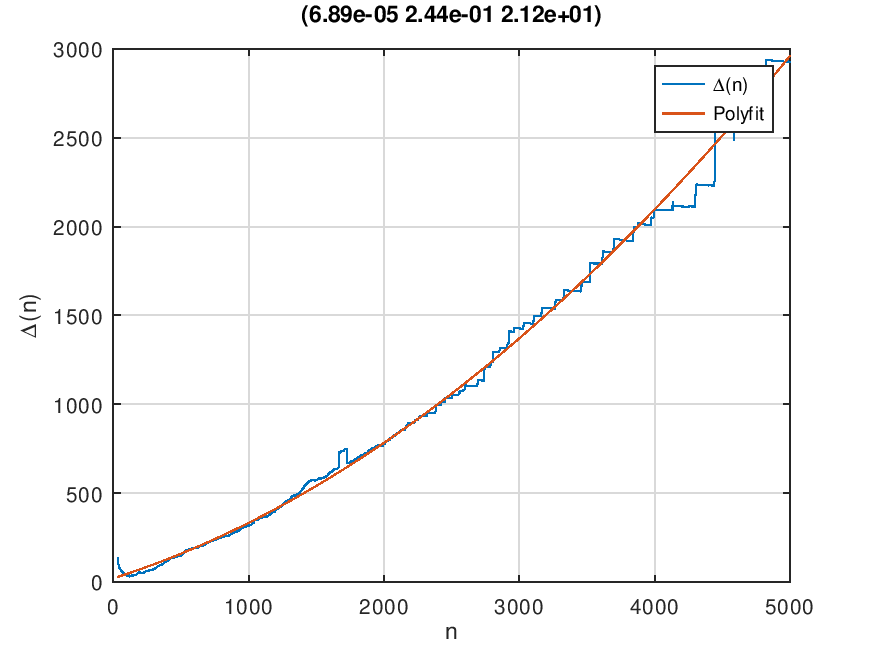}
    \caption{$\Delta(n)$.}
    \label{fig:delta}
\end{subfigure}
\hfill
\begin{subfigure}{0.495\textwidth}
    \includegraphics[width=.8\columnwidth]{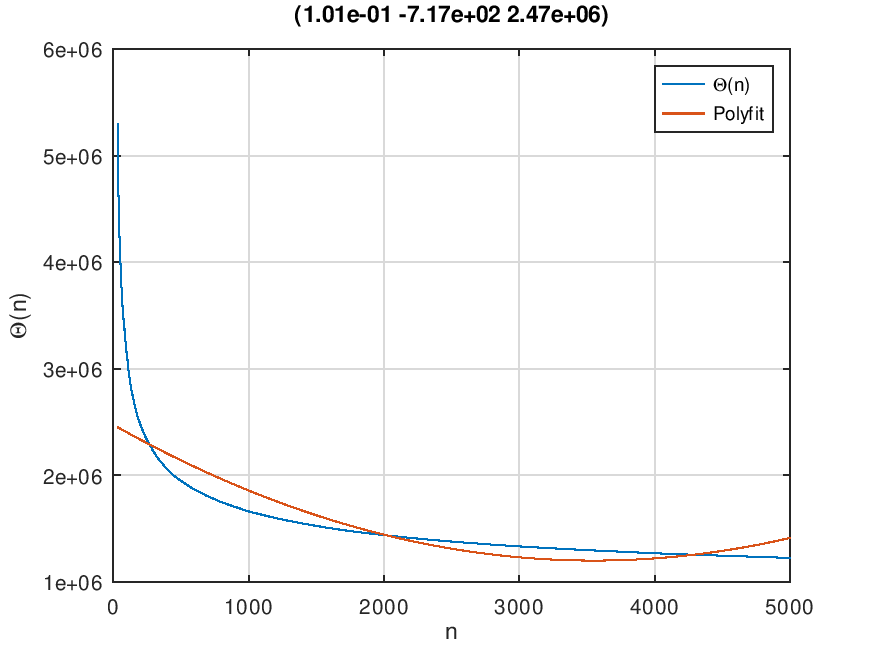}
    \caption{$\Theta(n)$.}
    \label{fig:theta}
\end{subfigure}
\caption{Second-order polynomial fit (Eq.~(\ref{eq:polyfit})) of (a) $\Delta(n)$ and (b) $\Theta(n)$. The fitted curves are shown in red, while the empirical curves derived from the LibriSpeech-100 corpus are shown in blue.}
\label{fig:delta-theta-polyfit}
\end{figure}

The $R^2$ metric (coefficient of determination) indicates the goodness of fit, with values closer to $1$ implying a better fit~\cite{montgomery2021introduction}.
The fitted $\Delta(n)$ and $\Theta(n)$ curves are shown in red in Figures~\ref{fig:delta} and~\ref{fig:theta}, respectively, while the empirical curves derived from the LibriSpeech-100 corpus are shown in blue.
As observed earlier, the second-order polynomial fit does not accurately capture $\Delta(n)$ for $n > 2500$, and similarly fails to model $\Theta(n)$ well for $n < 2500$.

Using Eq.~(\ref{eq:found_generic_n}), the minimizing value of $n$ is given by
\begin{equation}
n =
\frac{\tonep \alpha_3 - \alpha_1 - \done \alpha_2}
{2(\dtwo \alpha_2 + \ttwo \alpha_3)}
\label{eq:foundn}
\end{equation}
subject to the conditions that
\begin{equation}
2(\dtwo \alpha_2 + \ttwo \alpha_3) > 0
\label{eq:c1}
\end{equation}
and
\begin{equation}
\tonep \alpha_3 - \alpha_1 - \done \alpha_2 < 0.
\label{eq:c2}
\end{equation}
As noted earlier, the identified value of $n$ depends on the choice of $\alpha_{1,2,3}$ and can be computed only once these weights are specified \textit{a priori}.

\subsubsection{Finding \texorpdfstring{$n$}{n}}

State-of-the-art automatic speech recognition (ASR) systems commonly employ the Conformer encoder-decoder architecture.
The Conformer model implemented in the ESPNet toolkit~\cite{watanabe18_interspeech}, using the LibriSpeech-100 (low-resource) recipe, recommends an \SP-\unigram\ language model with $n=300$. 
We verified that multiple configurations of $\alpha_{1,2,3}$ map to the same heuristic value of $n$, which motivates the need for normalization (see Appendix~\ref{sec:n300}).

To determine an optimal value of $n$, greater control over the admissible range of $\alpha_{1,2,3}$ is required so that an optimal $n$ can be obtained via~\eqref{eq:foundn}. 
We address this next.

Observe that the ranges of the three terms $t_{1,2,3}$ vary significantly (Table~\ref{tab:terms_max_min}).
\begin{table}[!htb]
    \centering
    \begin{tabular}{|c||c|c|} \hline
    \multirow{2}{*}{Term} & \multicolumn{2}{c|}{Value}\\ \cline{2-3}
       & Min & Max\\ \hline
       ${t_1 = n}$  & $c_u$  & $\words_u$ \\
       ${t_2 = \Delta(n)}$ & $0$ & $f_c^+$ \\
       ${t_3 = \Theta(n)}$ & $0$ & $c$ \\ \hline
    \end{tabular}
    \caption{Range of the terms $t_{1,2,3}$. Here, $c_u$ denotes the number of unique characters, $\words_u$ the number of unique words, $f_c^+$ the frequency of the most frequent character, and $c$ the total number of characters.}
    \label{tab:terms_max_min}
\end{table}

From the corpus, we can directly compute the total number of words ($\words$), the number of unique words ($\words_u$), the total number of characters ($c$), the number of unique characters ($c_u$), and the frequency of the most frequent character ($f_c^+$). Hence, the minimum and maximum values of $t_{1,2,3}$ in Table~\ref{tab:terms_max_min} are known \textit{a priori}. 
For example, the term $t_1=n$ can take values in the range $[c_u, \words_u]$.

Normalizing $t_{1,2,3}$ using these bounds allows us to define a normalized cost function
\begin{equation}
     \Cnorm =
     \alpha'_1 \tnorm_1 +
     \alpha'_2 \tnorm_2 +
     \alpha'_3 \tnorm_3,
     \label{eq:mimimize_norm}
\end{equation}
where $\tnorm_1 = (t_1 - c_u)/(\words_u - c_u)$, $\tnorm_2 = t_2/f_c^+$, and $\tnorm_3 = t_3/c$. 
The weights $\alpha'_{1,2,3}$ now reflect the relative importance assigned to the normalized components.
Substituting for $\tnorm_{1,2,3}$, we obtain
\begin{equation}
     \Cnorm =
     \alpha'_1 \left ( \frac{n-c_u}{\words_u - c_u} \right )
     + \alpha'_2 \Delta^{norm}(n)
     + \alpha'_3 \Theta^{norm}(n),
     \label{eq:mimimize_new}
\end{equation}
where $\Delta^{norm}(n) = \Delta(n)/f_c^+$ and $\Theta^{norm}(n) = \Theta(n)/c$.

\begin{figure}[!htb]
    \centering
    \includegraphics[width=1\columnwidth]{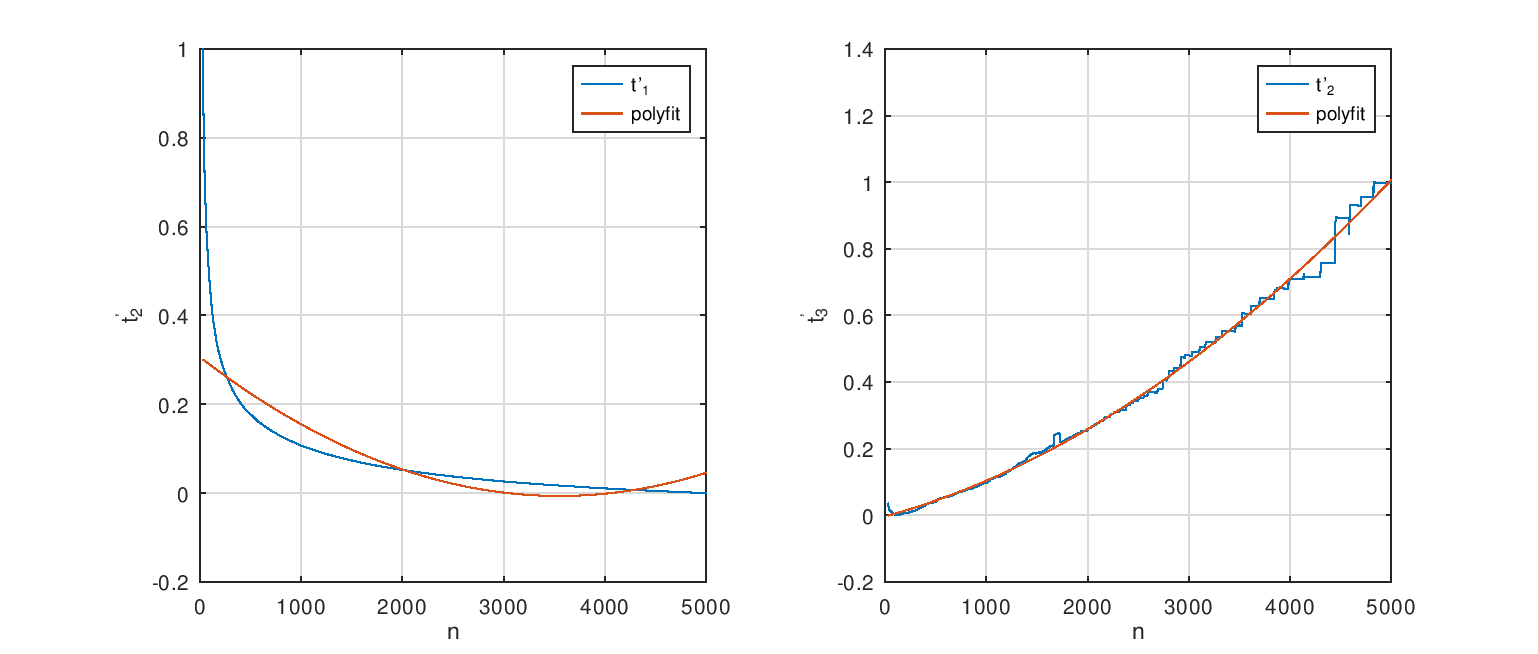}
    \caption{ $\Delta^{norm}(n)$ (right), $\Theta^{norm}(n)$ (left) shown in red are the result of second order polynomial fit (Eq (\ref{eq:polyfit_norm})) while $\Delta^{norm}(n)$ and $\Theta^{norm}(n)$ derived from LibriSpeech-100 corpus is shown as the blue curve. The y-axis is between $0$ and $1$ because of the normalization as seen in (\ref{eq:polyfit_norm}).
    }
    \label{fig:polyfit_norm}
\end{figure}

As before, we fit second-order polynomials to $\Delta^{norm}(n)$ and $\Theta^{norm}(n)$, yielding (see Fig.~\ref{fig:polyfit_norm})
\begin{eqnarray}
    \Delta^{norm}(n) &=& \ddtwo n^2  \ddone n + \ddzero \nonumber \\
    \Theta^{norm}(n) &=& \tttwo n^2 + \ttone n  \ttzero. \nonumber \\
    \label{eq:polyfit_norm}
\end{eqnarray}

The minimizing value of $n$ is then given by
\begin{equation}
n =
\frac{-(\ttone \alpha'_3 + \alpha'_1  \ddone \alpha'_2)}
{2(\ddtwo \alpha'_2 + \tttwo \alpha'_3)},
\label{eq:foundnn}
\end{equation}
subject to
\begin{equation}
2(\ddtwo \alpha'_2 + \tttwo \alpha'_3) > 0
\label{eq:generic_n_min_conditionn}
\end{equation}
and
\begin{equation}
(\alpha'_1  \ddone \alpha'_2 + \ttone \alpha'_3) < 0,
\label{eq:generic_n_num_conditionn}
\end{equation}
to ensure $n>0$.

\begin{algorithm}[!htb]
\caption{Find $n$ (using second order polynomial)} 
\label{algo:findn}
\begin{algorithmic}[1]
\STATE $\ddcoef{2}= \ddtwo$,
$\ddcoef{1}= \ddone$,
$\ddcoef{0}= \ddzero$
\STATE
$\ttcoef{2}= \tttwo$,
$\ttcoef{1}= \ttone$,
$\ttcoef{0}= \ttzero$ \hfill \COMMENT{As mentioned in (\ref{eq:polyfit_norm})}
\STATE
\STATE \textbf{Define the objective function, constraints, initialization:}
\STATE \hspace{1em} ${\tt obj}(\alpha'_1, \alpha'_2, \alpha'_3, \ttcoef{1}, \ttcoef{2}, \ddcoef{1}, \ddcoef{2})$
\STATE \hspace{2em} 
 $n = \frac{-( \ttcoef{1}* \alpha'_3 + \alpha'_1 +  \ddcoef{1} * \alpha'_2)}{(2*( \ddcoef{2} * \alpha'_2 +  \ttcoef{2}* \alpha'_3))}$ \label{algo:npoly_objective} \hfill \COMMENT{Equation (\ref{eq:foundnn})}
\STATE \hspace{2em} \textbf{return} $n$
\STATE \hspace{1em} ${\tt cons} \gets 2(\ddcoef{2}\alpha'_2 + \ttcoef{2}\alpha'_3) > 0, -(\ttcoef{1} \alpha'_3 +\alpha'_1 + \ddcoef{1} \alpha'_2) < 0$, $0 \le \alpha'_{1,2,3}\le 1$ and $\sum_{i=1}^3\alpha'_i = 1$ \label{algo:npoly_constraint} \hfill \COMMENT{Equations (\ref{eq:generic_n_min_conditionn}) and (\ref{eq:generic_n_num_conditionn})}
\STATE \hspace{1em} ${\tt init} \gets {\tt np.random.rand(3)} $ \label{algol:minimize_normalize}
\STATE
\STATE \textbf{Optimize:}
\STATE $result \gets {\tt minimize}({\tt obj}, {\tt init}, constraints={\tt cons}, method={\tt 'SLSQP'})$ \label{algo:npoly_solver}
\IF {$result.success$}
    \STATE $\alpha'_1, \alpha'_2, \alpha'_3 \gets result.x$ \label{algo:alphas}
\ELSE
    \STATE choose another ${\tt init}$ \hfill \COMMENT{not all initial values of $\alpha$'s result in a solution}
    \STATE Go to Line \#\ref{algol:minimize_normalize}
\ENDIF
\end{algorithmic}
\end{algorithm}

\begin{table}[t]
\caption{WERs (in \%; $\downarrow$ better) on LibriSpeech-100. ``avg'': average over ``clean'' and ``other''.}
\label{tab:results_poly}
\begin{center}
\resizebox{\columnwidth}{!}
{
\begin{tabular}{||c||cc|c||cc|c||} \hline
\multicolumn{7}{||c||}{\SP-\unigram} \\ \hline
\multirow{2}{*}{$n (\alpha'_{1,2,3})$}&\multicolumn{3}{c||}{dev} & \multicolumn{3}{c||}{test} \\ \cline{2-7}
 & clean & other & avg & clean & other & avg \\ \hline
300 (-)& 7.70 & 20.00 & 13.85 & 8.30 & 20.80 & 14.55 \\ 
382 (0.00, 0.39, 0.61)   &  7.80 & 20.10 & 13.95 & {\bf 8.00} & {\bf 20.70} & {\bf 14.35}\\ \hline 
\end{tabular}
}
\end{center}
\end{table}

\subsubsection{Performance on Speech Recognition}

As described earlier, our experiments use the LibriSpeech-100 corpus, consisting of 100 hours of read English speech with corresponding text transcriptions~\cite{train-clean-100}. 
The text transcripts are used to estimate the optimal number of tokens ($n$), and the effectiveness of the estimated vocabulary size is then validated by training ASR systems using the same tokens. 
We report recognition performance on the ``test-clean''~\cite{test-clean-100} and ``test-other''~\cite{test-other-100} evaluation sets (sentences: $2{,}939$; words: $52{,}343$).

We employ the state-of-the-art Conformer encoder–decoder architecture~\cite{gulati20_interspeech} for ASR. 
The Conformer model is implemented using the ESPNet toolkit~\cite{watanabe18_interspeech} with the LibriSpeech-100 (low-resource) training recipe. 
To ensure a fair comparison, we modify \emph{only} the number of sub-word units hyper-parameter in the original recipe while keeping all other model hyper-parameters unchanged.

The encoder consists of 12 layers and the decoder consists of 6 layers. 
The model dimension is 256 with 4 attention heads. 
Training is performed using the Adam optimizer~\cite{kingma2017adam} with $\beta_1 = 0.9$, $\beta_2 = 0.98$, and $\epsilon = 10^{-9}$, following the optimization strategy in~\cite{vas_attn}. 
A warm-up schedule of 25\,000 steps is used, and all models are trained for 100 epochs with a batch size of 64, consistent with the ESPNet recipe. 
All experiments are conducted using a single NVIDIA RTX 3090 GPU.

No language model is employed during decoding via shallow fusion. 
The sub-word units derived from the training text serve as the output labels for the ASR model. 
The input features consist of 80-dimensional log Mel spectrograms augmented with pitch, resulting in a total of 81-dimensional feature vectors. 
Additionally, standard data augmentation techniques are applied, including three-way speed perturbation~\cite{ko15_interspeech} with factors 0.9, 1.0, and 1.1, along with SpecAugment~\cite{park19e_interspeech}.

Table~\ref{tab:results_poly} summarizes the ASR performance in terms of word error rate (WER; lower is better) for the default ESPNet setting ($n=300$) and for the estimated optimal vocabulary size ($n^*=382$). 
While the overall performance for $n=300$ and $n^*=382$ is comparable, a modest improvement is observed for $n^*$ on the ``test-avg'' set (14.35 versus 14.55). 
For ``dev-avg'', the performance with $n^*=382$ (13.95) is close to that obtained with $n=300$ (13.85).

We emphasize that the commonly used value of $n=300$, to the best of our knowledge, lacks a principled justification in the literature. 
In contrast, this work proposes a formal, analytically grounded framework for estimating the vocabulary-size hyper-parameter, yielding $n^*=382$ for LibriSpeech-100.

However, we note that the second-order polynomial does not adequately fit $\Theta^{norm}(n)$, as discussed earlier and illustrated in Fig.~\ref{fig:polyfit_norm}. 
This observation motivates the exploration of a more expressive model by incorporating an exponential term alongside the second-order polynomial. 
We hypothesize that improved curve fitting of the empirical data can lead to a more accurate estimate of the optimal vocabulary size $n^*$.

\def\dcthree{6.8e-05}
\def\dctwo{2.47e-01}
\def\dcone{1.15e+03}
\def\dczero{-1.14e+03}
\def\tcthree{3.8e-02}
\def\tctwo{-3.12e+02}
\def\tcone{1.12e+08}
\def\tczero{-1.11e+08}

\subsection{Exponential and Second-Order Polynomial Fit}
\label{sec:second_order_plus_fit_expt}

As discussed in Section~\ref{sec:second_order_plus_fit}, we augment the second-order polynomial with an exponential term to model $\Delta(n)$ and $\Theta(n)$. 
The parameters of the resulting functions are estimated using the {\tt curve\_fit} routine from the {\tt scipy.optimize} module in the {\tt SciPy} library. 
The fitted models are given by
\begin{align}
    \Deltaexp(n) &= \dcthree *n^2 + \dctwo*n \nonumber \\ 
     &+ \dcone*\exp^{\left (\frac{1}{n} \right )} \dczero \nonumber \\
    \Thetaexp(n) &= \tcthree*n^2  \tctwo * n  \nonumber \\
    &+ \tcone *\exp^{\left (\frac{1}{n} \right )}  \tczero 
    \label{eq:polyfit_plus}
\end{align}
where the fitted parameters for $\Delta_{\mathrm{exp}}(n)$,
$\{\dccoef{3}, \dccoef{2}, \dccoef{1}, \dccoef{0}\}
= \{\dcthree, \dctwo, \dcone, \dczero\}$,
yield an $R^2$ value of $1.00$, and those for $\Theta_{\mathrm{exp}}(n)$,
$\{\tccoef{3}, \tccoef{2}, \tccoef{1}, \tccoef{0}\}
= \{\tcthree, \tctwo, \tcone, \tczero\}$,
yield an $R^2$ value of $0.99$.

\begin{figure}[!htb]
\centering
\begin{subfigure}{0.495\textwidth}
    \includegraphics[width=.8\columnwidth]{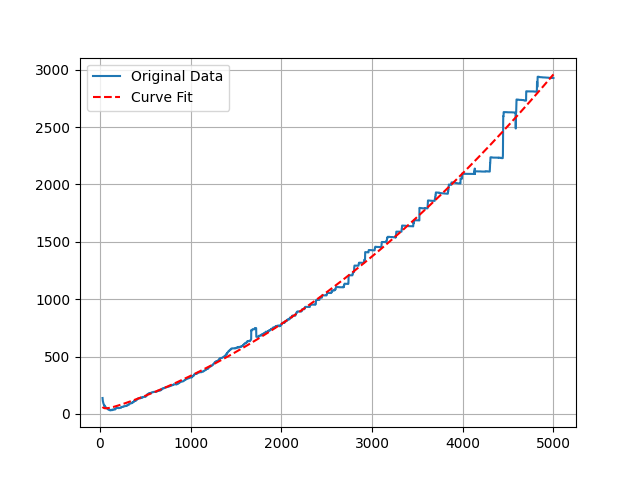}
    \caption{$\Deltaexp(n)$.}
    \label{fig:deltac}
\end{subfigure}
\hfill
\begin{subfigure}{0.495\textwidth}
    \includegraphics[width=.8\columnwidth]{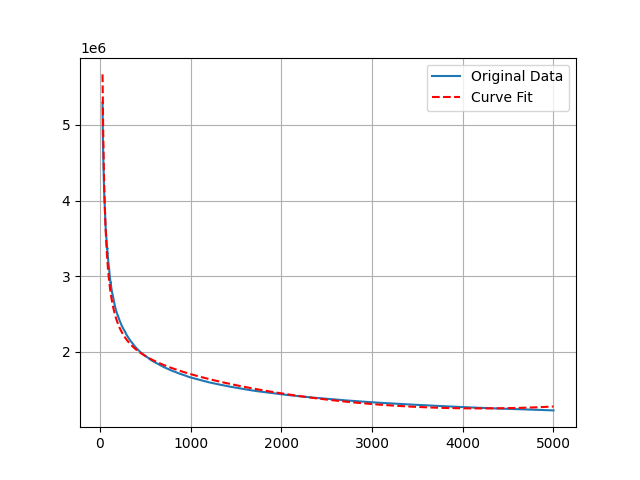}
    \caption{$\Thetaexp(n)$.}
    \label{fig:thetac}
\end{subfigure}
\caption{Exponential and second order polynomial to represent $\Delta(n)$ and $\Theta(n)$. Shown in Eq (\ref{eq:polyfit}).}
\label{fig:delta-theta-polyfit-plus}
\end{figure}

The improved goodness of fit is evident from the red curves in Fig.~\ref{fig:deltac} for $\Delta_{\mathrm{exp}}(n)$ and Fig.~\ref{fig:thetac} for $\Theta_{\mathrm{exp}}(n)$.
We hypothesize that this improved functional representation enables a more accurate estimation of the optimal vocabulary size $n$.

As before, solving the first-order optimality condition yields
\begin{eqnarray}
2n^3 \left (\dcthree*\alpha_2  + \tcthree*\alpha_3 \right ) &+&  \\
n^2 \left ( \alpha_1 + \dctwo * \alpha_2 + \tctwo * \alpha_3  \right )
 &-& \nonumber \\
 \exp^{\left (\frac{1}{n} \right )}
\left (\dcone * \alpha_2 + \tcone * \alpha_3 \right ) & = & 0 \nonumber
\label{eq:found_generic_cn_values}
\end{eqnarray}
subject to the second-order condition
\begin{eqnarray}
    \alpha_2 \left ( 2*\dcthree + 3* \dcone * \frac{\exp^{\left (\frac{1}{n} \right )}}{n^4} \right ) &+&  \\
    \alpha_3 \left ( 2*\tcthree + 3*\tcone *\frac{\exp^{\left (\frac{1}{n} \right )}}{n^4} \right ) &>& 0 \nonumber
    \label{eq:generic_cn_min_condition_values}
\end{eqnarray}
which ensures that the obtained solution corresponds to a minimum of the cost function.

Equations~\eqref{eq:found_generic_cn_values} and~\eqref{eq:generic_cn_min_condition_values} are obtained by substituting the fitted coefficients from~\eqref{eq:polyfit_plus} into the general expressions in~\eqref{eq:found_generic_cn} and~\eqref{eq:generic_cn_min_condition}, respectively.
We use the {\tt fsolve} function from the {\tt scipy.optimize} module to numerically solve~\eqref{eq:found_generic_cn_values} under the constraint~\eqref{eq:generic_cn_min_condition_values} (see Algorithm~\ref{algo:n_polyfit_plus}).

The solver is executed 25{,}000 times with different initializations of $\alpha_{1,2,3}$, while enforcing an additional constraint $n < 500$ (Algorithm~\ref{algo:n_polyfit_plus}, Line~\ref{algol:n500constraint}). 
In approximately 98\% of the runs, the solver converges to values of
\begin{align}
n^* \in [&57.49,\; 58.27,\; 58.46,\; 58.65,\; 58.84,\; \nonumber \\
&59.23,\; 59.42,\; 59.81,\; 60.00,\; 60.58],
\label{eq:final_n}
\end{align}
which lie in a narrow range around $n \approx 60$.

Notably, it was shown in~\cite{kopparapu2024cost} that the best-performing end-to-end ASR system achieved its lowest error rates when the vocabulary size was set to $n^* = 61$.
The values of $n^* \in (57, 61)$ obtained here are therefore consistent with, and closely match, that empirically optimal setting.

We emphasize that in~\cite{kopparapu2024cost}, the weights $\alpha_{1,2,3}$ were selected heuristically via grid search. 
In contrast, the present work demonstrates that the optimal vocabulary size can be estimated analytically by identifying a double-differentiable function that accurately fits $\Delta(n)$ and $\Theta(n)$ derived from the training corpus.
As shown in Table~\ref{tab:results_poly_plus}, the ASR system trained with $n^* = 61$ outperforms the commonly adopted ESPNet configuration with $n = 300$ on both ``dev-avg'' (13.20 versus 13.85) and ``test-avg'' (13.60 versus 14.55) subsets of LibriSpeech-100.

The proposed framework relies on identifying a suitable twice-differentiable functional form that provides a high-quality curve fit (i.e., high $R^2$ values) to the empirical $\Delta(n)$ and $\Theta(n)$ statistics. 
The dependence on the quality of this fit constitutes the primary limitation of the approach.

\begin{algorithm}
\caption{Find $n$ (modeled using second order polynomial and exponential term)}
\begin{algorithmic}[1]
    \STATE \textbf{Step 1: Generate Random $\alpha_1, \alpha_2, \alpha_3$}
    \STATE Generate 3 random values $r_1, r_2, r_3$ between 0 and 1
    \STATE Normalize: $\alpha_1 = \frac{r_1}{r_1 + r_2 + r_3}, \alpha_2 = \frac{r_2}{r_1 + r_2 + r_3}, \alpha_3 = \frac{r_3}{r_1 + r_2 + r_3}$
\STATE
    \STATE \textbf{Step 2: Define Equation (\ref{eq:found_generic_cn_values}})
    \STATE Define $f(n, \alpha_1, \alpha_2, \alpha_3) = 2n^3(6.8e-05\alpha_2 + 3.8e-02\alpha_3) + n^2(\alpha_1 + 2.47e-01\alpha_2 - 3.12e+02\alpha_3) - \exp(1/n)(1.15e+03\alpha_2 + 1.12e+08\alpha_3)$
\STATE
    \STATE \textbf{Step 3: Define Constraint Equation (\ref{eq:generic_cn_min_condition_values})}
    \STATE Define $g(n, \alpha_2, \alpha_3) = \alpha_2(2 \times 6.8e-05 + 3 \times 1.15e+03 \times (\exp(1/n) / n^4)) + \alpha_3(2 \times 3.8e-02 + 3 \times 1.12e+08 \times (\exp(1/n) / n^4))$
\STATE
    \STATE \textbf{Step 4: Set Initial Guess for $n$}
    \STATE Set $n\_initial = 1.0$
\STATE
    \STATE \textbf{Step 5: Solve the Equation}
    \IF{using \texttt{fsolve}}
        \STATE Solve $f(n, \alpha_1, \alpha_2, \alpha_3) = 0$ with constraint $g(n, \alpha_2, \alpha_3) > 0$
    \ELSE
        \STATE Minimize $|f(n, \alpha_1, \alpha_2, \alpha_3)|$ with bounds on $n < 500$ \label{algol:n500constraint}
    \ENDIF
\STATE
    \STATE \textbf{Step 6: Print the Results}
    \STATE Print values of $\alpha_1$, $\alpha_2$, $\alpha_3$, and $n$
    
\end{algorithmic}
\label{algo:n_polyfit_plus}
\end{algorithm}

\begin{table}[t]
\caption{WERs (in \%; $\downarrow$ better) on LibriSpeech-100. ``avg'': average over ``clean'' and ``other''.}
\label{tab:results_poly_plus}
\begin{center}
{
\begin{tabular}{||c||cc|c||cc|c||} \hline
\multicolumn{7}{||c||}{\SP-\unigram} \\ \hline
\multirow{2}{*}{$n$}&\multicolumn{3}{c||}{dev} & \multicolumn{3}{c||}{test} \\ \cline{2-7}
 & clean & other & avg & clean & other & avg \\ \hline
300 & 7.70 & 20.00 & 13.85 & 8.30 & 20.80 & 14.55 \\ 
{61} \cite{kopparapu2024cost}& {\bf 7.20} & {\bf 19.20} & {\bf 13.20} & {\bf 7.70} & {\bf 19.60} & {\bf 13.60}\\ 
\hline
\end{tabular}
}
\end{center}
\end{table}

\section{Conclusions}
\label{sec:conclude}

End-to-end deep learning architectures have become the dominant paradigm across several application domains, including automatic speech recognition (ASR). These models rely on a set of hyper-parameters whose careful selection is critical for achieving optimal performance. In practice, however, hyper-parameters are often chosen heuristically or through extensive empirical searches over large configuration spaces. This reliance on heuristics remains a common limitation of most end-to-end systems, including ASR.

Unlike hybrid ASR systems, where the vocabulary is fixed and derived from the phonetic structure of the target language, end-to-end ASR systems introduce vocabulary size as an explicit and influential hyper-parameter. Popular frameworks such as ESPNet provide predefined vocabulary sizes as part of their training recipes. However, these choices are typically not accompanied by an explanation of the underlying rationale or a principled method for their derivation. To the best of our knowledge, systematic studies examining the impact of vocabulary size on end-to-end ASR performance have been limited until recent work demonstrated its measurable influence~\cite{kopparapu2024cost}.

In this paper, we introduce a mathematical formalism that brings interpretability and analytical rigor to the process of selecting the vocabulary size hyper-parameter, moving beyond heuristic choices prevalent in existing literature. Leveraging the first- and second-derivative tests from classical calculus, we model corpus-dependent statistics using (i) a second-order polynomial and (ii) a second-order polynomial augmented with an exponential term. These functional forms ensure the existence of well-defined first- and second-order derivatives, enabling analytical characterization of optimality conditions.

The resulting equations are solved numerically to estimate the optimal vocabulary size for a given training corpus. Using the LibriSpeech-100 dataset, we observe that incorporating an exponential term significantly improves the modeling of corpus statistics compared to a pure second-order polynomial. This improved fit leads to a more accurate estimate of the optimal vocabulary size, which is further validated through end-to-end ASR experiments. Our results demonstrate that the proposed framework can effectively identify a principled vocabulary size that yields improved recognition performance relative to commonly adopted heuristic choices.

Overall, this work provides a formal and explainable methodology for estimating the vocabulary size hyper-parameter in end-to-end ASR systems, addressing an important gap in current practice.

As a limitation, the applicability of the proposed framework depends on the ability to accurately fit twice-differentiable functions to the corpus-derived statistics $\Delta(n)$ and $\Theta(n)$. The quality of the estimated vocabulary size is therefore contingent on the fidelity of this curve-fitting process, which represents a known constraint of the proposed approach.

\section*{Acknowledgments}
Would like to acknowledge the experiments conducted by Dr Ashish Panda which have been reported in Table \ref{tab:results_poly} and Table \ref{tab:results_poly_plus}.

\section*{Statement on GenAI Usage}

The formulation of the problem, the mathematical modeling, and the analytical derivation of the optimal vocabulary size are entirely the work of the author. Generative AI (Copilot) was used solely as an assistive tool to improve the clarity, organization, and readability of the manuscript. In particular, Copilot provided suggestions related to phrasing, flow, and structural organization—for example, recommending that material initially included in the main text be moved to the Appendix to improve the overall presentation. No scientific content, analysis, results, or conclusions were generated by the AI.

\bibliography{specom-bib}

\appendices

\section{Exponential and Second order Polynomial Fit}
\label{sec:second_order_plus_fit_a}
As mentioned, $\Deltaexp(n)$ (\ref{eq:deltac}) and $\Thetaexp(n)$ (\ref{eq:thetac}) represent  the second order polynomial with an exponential term. 

We can compute the first and second derivative as  
\begin{eqnarray}
\Deltaexp'(n) &=& 2\dccoef{3}n + \dccoef{2} - \dccoef{1} 
\frac{\exp^{\left (\frac{1}{n} \right )}}{n^2}, \nonumber \\ 
\Thetaexp'(n) &=& 2\tccoef{3}n + \tccoef{2} - \tccoef{1} 
\frac{\exp^{\left (\frac{1}{n} \right )}}{n^2} \nonumber
\end{eqnarray} 
\begin{eqnarray} 
\Deltaexp''(n) &=& 2\dccoef{3} + 3 \dccoef{1} \frac{\exp^{\left (\frac{1}{n} \right )}}{n^4},  \nonumber \\
\Thetaexp''(n) &=& 2\tccoef{3} + 3 \tccoef{1} \frac{\exp^{\left (\frac{1}{n} \right )}}{n^4} \nonumber.
\end{eqnarray}
Substituting 
in (\ref{eq:derivative_min}) we get 
\begin{align}
\alpha_1 
+ \alpha_2
\left(
2\dccoef{3}n + \dccoef{2}
- \dccoef{1}\frac{\exp^{\left(\frac{1}{n}\right)}}{n^2}
\right) \nonumber\\
+ \alpha_3
\left(
2\tccoef{3}n + \tccoef{2}
- \tccoef{1}\frac{\exp^{\left(\frac{1}{n}\right)}}{n^2}
\right) = 0 \nonumber\\
\implies\;
\alpha_1 
+ \alpha_2 \dccoef{2}
+ \alpha_3 \tccoef{2}
+ 2n\left(\alpha_2 \dccoef{3} + \alpha_3 \tccoef{3}\right) \nonumber\\
-
\frac{\exp^{\left(\frac{1}{n}\right)}}{n^2}
\left(\alpha_2 \dccoef{1} + \alpha_3 \tccoef{1}\right) = 0 \nonumber\\
\implies\;
2n^3\left(\alpha_2 \dccoef{3} + \alpha_3 \tccoef{3}\right)
+ n^2\left(
\alpha_1 + \alpha_2 \dccoef{2} + \alpha_3 \tccoef{2}
\right) \nonumber\\
-
\exp^{\left(\frac{1}{n}\right)}
\left(\alpha_2 \dccoef{1} + \alpha_3 \tccoef{1}\right) = 0
\label{eq:found_generic_cn}
\end{align}
a cubic equation. 
To find the roots of 
(\ref{eq:found_generic_cn}), we can use numerical method like Newton-Raphson or other root-finding algorithms, under the constraint (\ref{eq:generic_cn_min_condition}).


\section{Reverse Solving for \texorpdfstring{$\alpha$}{alpha} at \texorpdfstring{$n=300$}{n=300}}
\label{sec:n300}
As in Algorithm \ref{algo:n300} we use  the Sequential Least Squares Quadratic Programming (SLSQP) method  to find the value of $\alpha_{1,2,3}$ 
which satisfies $n=300$, a value \textit{heuristically} chosen in ESPNet recipe. 
Specifically, we use 
nonlinear solvers from the {\tt scipy.optimize} library in Python with different initial guesses of $\alpha_{1,2,3}$ (column 1, Table \ref{tab:n-300}). Table \ref{tab:n-300} shows a sample set of $\alpha_{1,2,3}$'s which results in  $n \approx 300$ 
when (\ref{eq:foundn}) is solved. As can be seen, different initial values of $\alpha_{1,2,3}$'s (Line \ref{algol:initial}, Algorithm \ref{algo:n300}) result in values of  $\alpha_{1,2,3}$'s (Line \ref{algol:final}, Algorithm \ref{algo:n300}) which  satisfy $n \approx 300$ (Line \ref{algol:n300}, Algorithm \ref{algo:n300}). Table \ref{tab:n-300} shows a list of $\alpha_{1,2,3}$'s which result in $n=300$. 
While different values of $\alpha_{1,2,3}$'s (example $(23.01, -31.56, 0.02)$ and $(263571.00, -361185.85, 246.75)$) are valid 
it can be seen that 
the ratios  $\left (\frac{\alpha_1}{\alpha_3}\right )$, $\left (\frac{\alpha_2}{\alpha_3}\right )$, and $\left (\frac{\alpha_3}{\alpha_3}\right )$ 
are quite close to each other (see Table \ref{tab:n-300}).
\begin{table}[!htb]
    \centering
    \resizebox{\columnwidth}{!}{
        \begin{tabular}{|c||c|c|c||c|c|c|} \hline
        \multicolumn{7}{|c|}{$n=300$} \\ \hline
    Initial ($\alpha_1,\alpha_2,\alpha_3$) & $\alpha_1$ & $\alpha_2$ & $\alpha_3$ 
    & $\left (\frac{\alpha_1}{\alpha_3}\right )$ 
    & $\left (\frac{\alpha_2}{\alpha_3}\right )$ 
    & $\left (\frac{\alpha_3}{\alpha_3}\right )$ \\ \hline
       (0.26, 0.02, 0.91) & 23.01 & -31.56  & 0.02 & 1150.5&-1578.0&1 \\
       (0.41, 0.67, 0.71) & 33.29 & -45.65  & 0.03 & 1109.7&-1521.7&1 \\ 
       (0.57, 0.81, 0.47) & 431.6 & -591.96  & 0.40 & 1079.0&-1479.9&1 \\
       (0.26, 0.81, 0.42) & 624.06 & -855.92  & 0.58 & 1076.0&-1475.7&1 \\
       (0.25, 0.29, 0.24) & 22549.10 & -30908.29  & 21.11 & 1068.2&-1464.2&1\\
       (0.05, 0.81, 0.89) & 93058.70 & -127609.68  & 87.08 & 1068.7&-1465.4&1\\
       (0.63, 0.30, 0.82) & 263571.00 & -361185.85  & 246.75 & 1068.2&-1463.8&1\\
       \hline
    \end{tabular}
    }
    \caption{Values of $\alpha_{1,2,3}$ which result in $n=300$ (\ref{eq:derivative_min}). 
    The left column is the initial values to Algorithm \ref{algo:n300}.}
    \label{tab:n-300}
\end{table}
\begin{algorithm}[!htb]
\caption{Solving for $\alpha_{1,2,3}$ for $n=300$ (value heuristically chosen in ESPNet recipe for LibriSpeech-100 corpus)}
\label{algo:n300}
\begin{algorithmic}[1]
\STATE \textbf{Define the objective function, constraints, initialization:}
\STATE \hspace{1em} ${\tt obj}(\alpha_1, \alpha_2, \alpha_3)$
\STATE \hspace{2em} 
$n \gets \frac{716.87 \times \alpha_3 - \alpha_1 - 0.24 \times \alpha_2}{1.38 \times 10^{-4} \times \alpha_2 + 0.202 \times \alpha_3}$ \label{algol:n300} \hfill \COMMENT{Equation (\ref{eq:foundn})}
\STATE \hspace{2em} \textbf{return} $(n - 300)^2$ \hfill \COMMENT{Returns $n$ close to $300$.}

\STATE \hspace{1em} ${\tt cons} \gets 2(\dtwo\alpha_2 + \ttwo\alpha_3) > 0, \tonep \alpha_3 -\alpha_1 - \done\alpha_2 < 0$ 
\STATE{\hfill \COMMENT{Equations (\ref{eq:c1}), (\ref{eq:c2})}}
\STATE \hspace{1em} ${\tt init} \gets {\tt np.random.rand(3)} $ \label{algol:initial} 
\hfill \COMMENT{Initializing $\alpha$'s}

\STATE{}

\STATE \textbf{Optimize:}
\STATE $result \gets {\tt minimize}({\tt obj}, {\tt init}, constraints={\tt cons}, method={\tt 'SLSQP'})$ \label{algol:minimize}

\IF {$result.success$} 
    \STATE $\alpha_1, \alpha_2, \alpha_3 \gets result.x$ \label{algol:final}
\ELSE
    \STATE choose another ${\tt init}$ \hfill \COMMENT{not all initial values of $\alpha$'s result in a solution}
    \STATE Go to Line \#\ref{algol:minimize}
\ENDIF
\end{algorithmic}
\end{algorithm}

\end{document}